\documentclass[10pt,twocolumn,letterpaper]{article}

\usepackage{iccv}
\usepackage{times}
\usepackage{epsfig}
\usepackage{graphicx}
\usepackage{amsmath}
\usepackage{amssymb}
\usepackage{nicefrac}
\usepackage{multirow}
\usepackage{rotating}
\usepackage{adjustbox}
\usepackage[pagebackref=true,breaklinks=true,letterpaper=true,colorlinks,bookmarks=false]{hyperref}

\iccvfinalcopy 
\def\iccvPaperID{} 

\begin{document}

\title{Multi-View Stereo with Single-View Semantic Mesh Refinement}

\author{Andrea Romanoni \hspace{0.35cm}  Marco Ciccone \hspace{0.35cm} Francesco Visin \hspace{0.35cm} Matteo Matteucci\\[1ex] 
Politecnico di Milano, Italy\\[1ex] 
\{andrea.romanoni, marco.ciccone, francesco.visin, matteo.matteucci\}@polimi.it
}

\maketitle

\begin{abstract}
While 3D reconstruction is a well-established and widely explored research topic, semantic 3D reconstruction has only recently witnessed an increasing share of attention from the
Computer Vision community. 
Semantic annotations allow in fact to enforce strong class-dependent priors, as planarity for ground and walls, which can be exploited to refine the reconstruction often resulting in non-trivial performance improvements.
State-of-the art methods propose volumetric approaches to fuse RGB image data with semantic labels; even if successful, they do not scale well and fail to output high resolution meshes.
In this paper we propose a novel method to refine both the geometry and the semantic labeling of a given mesh.
We refine the mesh geometry by applying a variational method that optimizes a composite energy made of a state-of-the-art pairwise photo-metric term and a single-view term that models the semantic consistency between the labels of the 3D mesh and those of the segmented images.
We also update the semantic labeling through a novel Markov Random Field (MRF) formulation that, together with the classical data and smoothness terms, takes into account class-specific priors estimated directly from the annotated mesh. This is in contrast to state-of-the-art methods that are typically based on handcrafted or learned priors. We are the first, jointly with the very recent and seminal work of \cite{blaha2017semantically}, to propose the use of semantics inside a mesh refinement framework.
Differently from \cite{blaha2017semantically}, which adopts a more classical pairwise comparison to estimate the flow of the mesh, we apply a single-view comparison between the semantically annotated image and the current 3D mesh labels; this improves the robustness in case of noisy segmentations.
\end{abstract}

\section{Introduction}
Modeling a scene from a set of images has been a long-standing and deeply explored problem for the Computer Vision community.
The goal is to build an accurate 3D model of the environment basing on the implicit tridimensional data contained in a set of 2D images. 
These methods can be useful to digitalize architectural heritage, reconstruct maps of cities or, in general, for scene understanding.

Most dense 3D reconstruction algorithms consider only grayscale or color images, but thanks to the advancements in semantic image segmentation \cite{benbouzid2012multiboost,zheng2015conditional,chen2016deeplab,visin2016reseg}, novel 3D reconstruction approaches that leverage semantic information have been proposed \cite{sengupta2013urban,hane2013joint,savinov2015discrete,kundu2014joint,blaha2016large}.

\begin{figure}[t]
\centering
\setlength{\tabcolsep}{2px}
\begin{tabular}{cc}
\includegraphics[width=0.48\columnwidth]{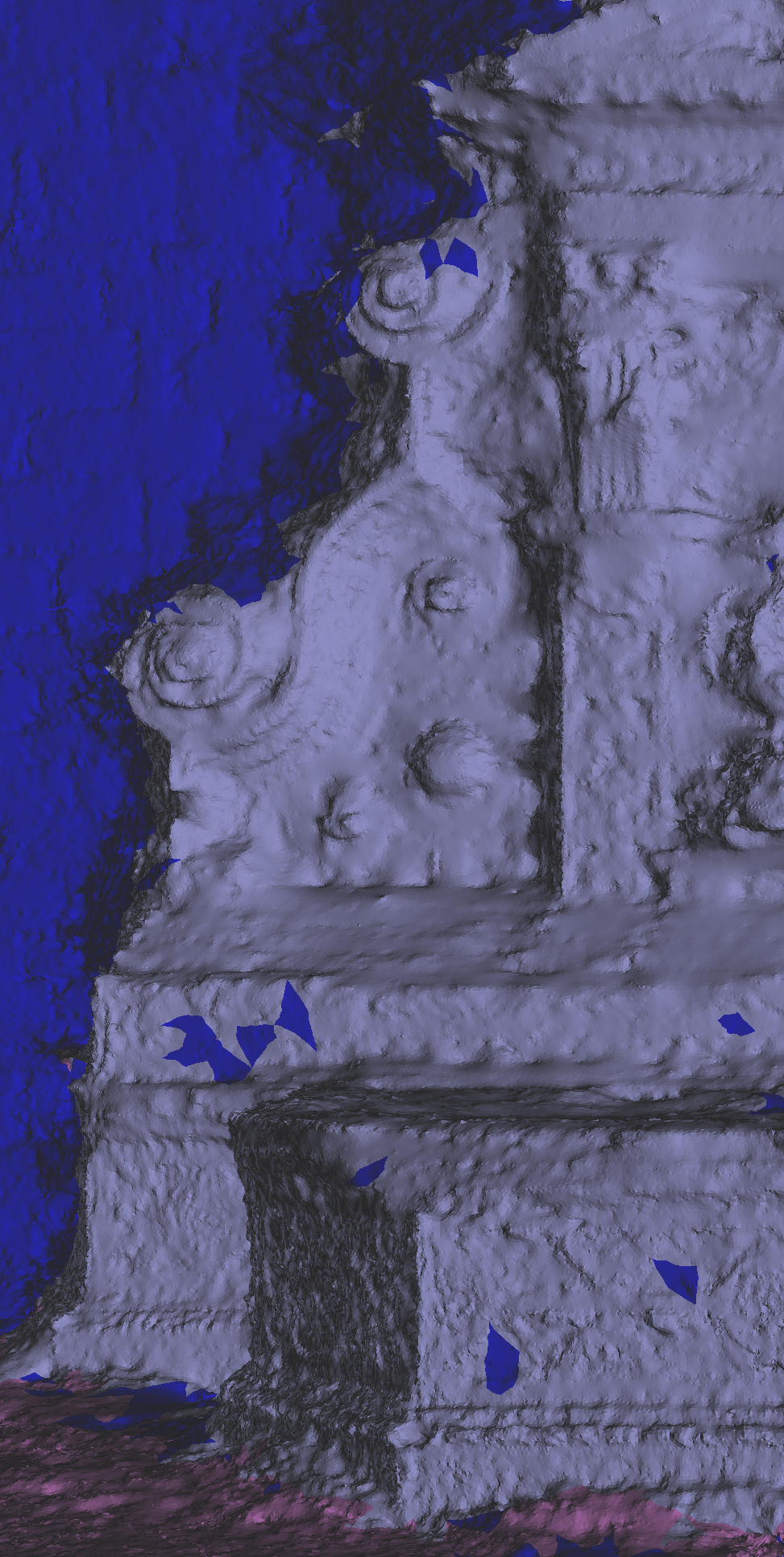}&
\includegraphics[width=0.48\columnwidth]{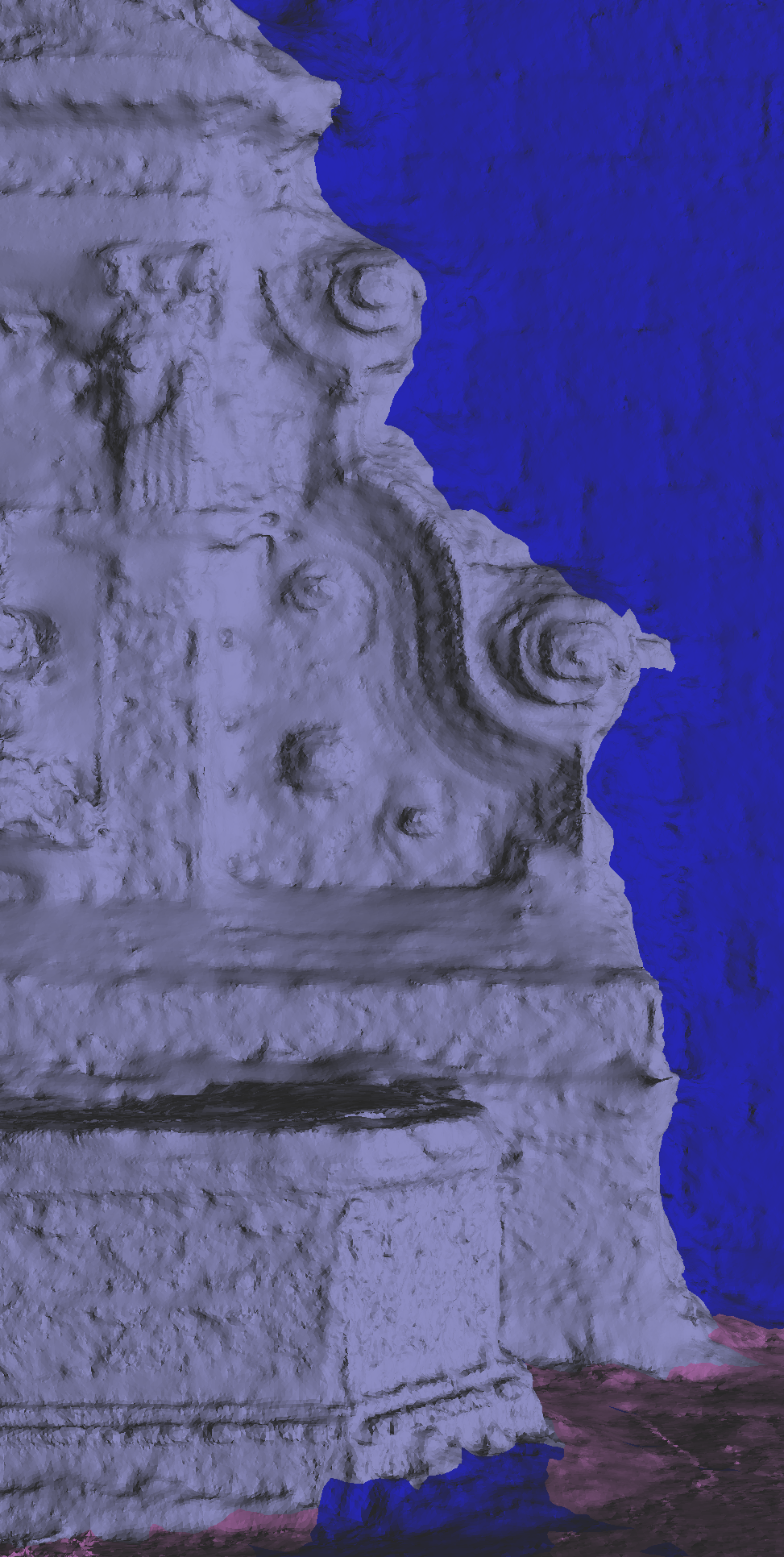}\\
\end{tabular}
\caption{The reconstruction of the fountain-p11 dataset \cite{strecha2008} without  (left) and with the semantic refinement (right) proposed in this paper}
\label{fig:fountain}
\end{figure}

State-of-the-art semantic dense 3D reconstruction algorithms fuse images and semantic labels in a volumetric voxel-based representation, improving the accuracy thanks to strong class-dependent priors (learned or handcrafted), e.g.,  planarity of the ground or perpendicularity between ground and walls.
These volumetric methods often allow to obtain impressive results, but they usually require a huge amount of memory. 
Only recently some effort has been put into solving this issue, for instance, through submaps \cite{cherabier2016multi} or multi-grid \cite{blaha2016large}.
Very recently a mesh refinement guided by semantic has been proposed in \cite{blaha2017semantically}: the authors update the reconstruction by minimizing the reprojection error between pairs of segmented images.
Differently from \cite{blaha2017semantically}, our work compares each segmented image against the labels fused into the 3D mesh. This is much more robust to noise and errors in the images segmentations, as we show in Section \ref{subsec:eth} and support with an in-depth experimentation and discussion.

The paper is structured as follows: Section \ref{sec:related_works} presents an overview of the state-of-the-art of several topics involved in the proposed system; these are then discussed in detail in the context of the proposed method in Section \ref{sec:method}. The experimental settings and results are presented in Section \ref{sec:exp} and discussed in Section \ref{sec:concl}.

\section{Related works}
\label{sec:related_works}
The method proposed in this paper crosses a variety of topics, namely classical and semantic volumetric reconstruction, photometric mesh refinement and mesh labeling. 
Here we review some of the most relevant works in those fields.
\paragraph{Volumetric 3D Reconstruction}
Volumetric 3D reconstruction represents the most widespread method to recover the 3D shape of an environment captured by a set of images.
These methods build a set of visibility rays either from Structure from Motion, as in \cite{litvinov_lhiuller14,romanoni15b}, or depth maps, as in \cite{pollefeys_et_al_08,newcombe2011dtam}. After the space is partitioned, these rays are used to classify the parts as being free space or matter. 
The boundary between free space and matter constitutes the final 3D model. 
Depending on how the space is discretized, volumetric algorithms are classified as voxel-based \cite{vogiatzis2005multi,steinbrucker2014volumetric} or tetrahedra-based \cite{tola12,vu2011large,litvinov_lhiuller14,romanoni16}.
The former trivially represent the space as a 3D grid.
Despite their simplicity, these approaches often lead to remarkable results; however their scalability is quite limited due to the inefficient use of space that does not take into account the significant sparsity of the elements that should be modeled.
Many attempt have been proposed to overcome this issue, e.g., by means of voxel hashing \cite{schops20153d}, but a convincing solution to the shortcomings of voxel-based methods seems to be still lacking.

On the other hand, tetrahedra-based approaches subdivide the space in tetrahedra via Delaunay triangulation: these methods build upon the points coming from Structure from motion or depth maps fusion and can adapt automatically to the different densities of the points in the space.
As opposed to voxel-based methods, tetrahedra-based approaches are scalable and can be very effective in a wide variety of scenarios; however, since they tend to restrict the model to those parts of the space occupied by the points, in some cases they can make it hard to define priors on the non-visible part of the scene (e.g., walls behind cars), since they might not be modeled.

\paragraph{Semantic Reconstruction}
A recent trend in reconstruction methods has been to embed semantic information to improve the consistency and the coherence of the produced 3D model \cite{kundu2014joint,hane2013joint}.
Usually these methods rely on voxels representation and estimate the 3D labeled model by enriching each camera-to-point viewing ray with semantic labels; these are then typically used to replace the ``matter'' label of the classical method.
The optimization process that leads to the final 3D reconstruction builds on class-specific priors, such as planarity for the walls or ground.
Being voxel-based, these approaches lack scalability: the authors of \cite{cherabier2016multi} tackle this issue via submaps reconstruction and by limiting the number of labels taken into account during the reconstruction of a single submap, while \cite{blaha2016large} adopts  multi-grids to avoid covering empty space with useless voxels.

Cabezas \textit{et al}. \cite{cabezas2015semantically} propose a semantic reconstruction algorithm that directly relies on mesh representation and fuses the data from aerial images, LiDAR and Open Street Map. Although proposing an interesting approach, such rich data is usually not available in a wide variety of applications, including the ones typical addressed in classical Computer Vision scenarios.
For a more detailed overview of semantic 3D reconstruction algorithms we refer the reader to \cite{hane2016overview}.

\paragraph{Photometric mesh refinement}
The approaches described so far extract the 3D model of the scene from a volumetric representation of it.
In some cases these models lack details and resolution, especially due to the scalability issue mentioned before.
Some works presented in the literature bootstrap from a low resolution mesh and refine it via variational methods \cite{yezzi2003stereoscopic,gargallo2007minimizing,pons2007multi,vu_et_al_2012,delaunoy_et_al_08}. 
Early approaches \cite{yezzi2003stereoscopic,gargallo2007minimizing} describe the surface as a continuous entity in $\mathit{R}^3$, minimize the pairwise photometric reprojection error among the cameras and finally discretize the optimized surface as a mesh.
More recently, some authors
 \cite{pons2007multi,vu_et_al_2012,delaunoy_et_al_08} proposed a few more effective methods that compute directly the discrete gradient that minimizes the reprojection error of each vertex in the mesh.
By relying on these methods Delaunoy and Pollefeys \cite{delaunoy2014photometric} proposed to couple the mesh refinement with the camera pose optimization.
Li \textit{et al.} \cite{li2016efficient} further improved the scalability of these methods by noticing that although mesh refinement algorithms usually increase the resolution of the whole mesh while minimizing the reprojection error, in some regions such as the flat ones there is no need for high vertex density. 
To avoid redundancy, \cite{li2016efficient} refine only the regions that produce a significant reduction of the gradient.

\paragraph{Mesh labeling}
Mesh labeling is usually modeled as a Markov Random Field (MRF) with a data term that describes the probability that a facet belongs to a certain class, and a smoothness term that penalizes frequent changes in the labeling along the mesh. 
Some approaches as \cite{verdie2015lod} rely on handcrafted priors that define relationships among the labels basing on their 3D position and orientation with respect to the neighbors.
Other methods add instead priors learned from data, such as \cite{valentin2013mesh,rouhani2017semantic}.

\subsection{Semantic mesh refinement}
\label{subsec:eth}
The very recent work presented in \cite{blaha2017semantically} exhibits some similarities with what we propose in this paper. As in our case, the authors propose a refinement algorithm that extends \cite{vu_et_al_2012} by leveraging semantic annotations.
In \cite{blaha2017semantically} the reprojection error between pairs of views is minimized in the same fashion as \cite{vu_et_al_2012}, although instead of using just RBG images they also use pairwise masks for each label taken into account by the semantic classifier.
The authors proved that this approach is effective and actually improves the photometric only refinement.
However, we show that in presence of noisy or wrong classification their method lacks robustness (see Section \ref{subsec:twoviews}) and we propose an alternative that does no suffer from this problem.
Secondly, although also the authors of \cite{blaha2017semantically} update the labels of the 3D mesh with a MRF with a data term, a smoothness term and handcrafted geometric priors, we propose a simpler data term that makes the refinement much less expensive in terms of computation and a term computed from the reconstructed labeled mesh that encourages the facets with one label to have similar distribution to the input mesh facets with the same label.

\section{Proposed method}
\label{sec:method}
\begin{figure}[t]
  \centering
  \includegraphics[width=0.995\columnwidth]{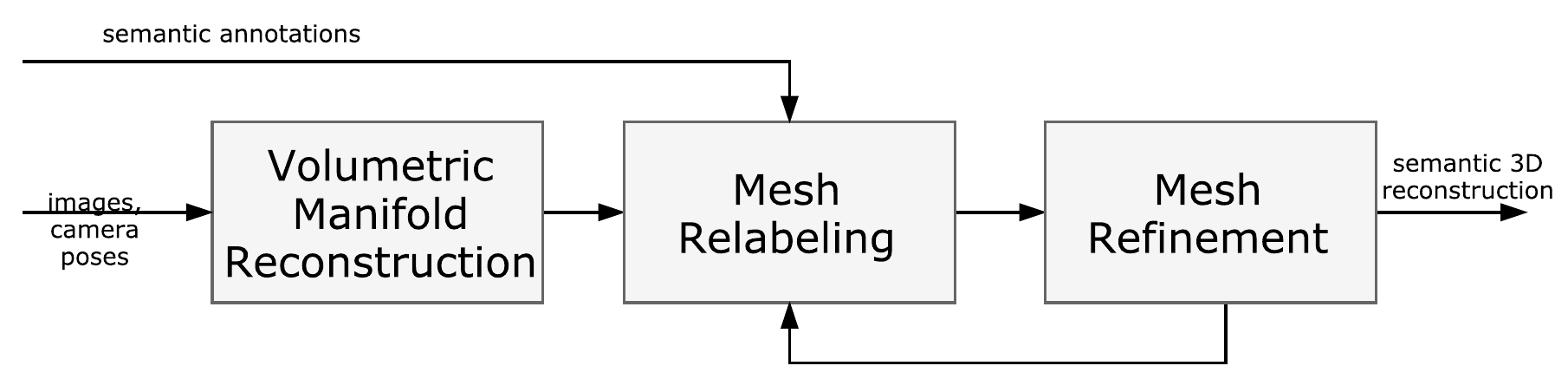}
  \caption{Architecture of the proposed system}
  \label{fig:architecture}
\end{figure}

The method we propose in this paper refines a labeled 3D mesh through a variational surface evolution framework: we alternate between the photo-consistent and semantic mesh refinement and the label update according to (Figure \ref{fig:architecture}).

The initialization of our method is the 3D mesh estimated and labeled by the modified version of \cite{romanoni15b}.
The volumetric method proposed in \cite{romanoni15b} estimates a point cloud from the images and discretizes the space through a Delaunay triangulation, initializing all the tetrahedra as \textit{matter}, i.e., with $0$ weight.
It then casts all camera-to-point rays and increases the weight of the traversed tetrahedra; finally it estimates the manifold surface that contains the highest number free space tetrahedra, i.e., those with weight above a fixed threshold.

To take into account the semantic labels associated to the image pixels, and in turn to the camera-to-point rays, in our version a tetrahedron has one weight associated to the free space label and one weight associated to each new semantic label. 
For each ray from camera $C$ to point $P$ associated to label $l$, we increase the free space weight of the tetrahedra between $C$ and $P$, as in the original case, then we increase the $l$ weight of the tetrahedra that, following the ray direction, are just behind (below a fixed distance) the point $P$, similarly to \cite{savinov2015discrete}.
Each tetrahedra is classified accordingly to the label with higher weight and the manifold is estimated as in the original version, but each triangle of the output mesh has now the label of the tetrahedron they belong to.

\subsection{Label smoothing and update}
\label{subsec:label}
In two cases we need to update the labeling of the 3D mesh: 1) after the initial mesh computation and 2) after the semantic segmentation.
After the volumetric reconstruction previously described, the labels of the initial mesh are prone to noise and, even if they collect more evidences for the same 3D point across the 2D set of semantically annotated images, sometimes they reflect the errors of the 2D classifier.
After the refinement process  the shape of the model changes, i.e., each facet change its orientation and positioning, therefore we update the labels to take into account the modifications of the value of the priors; moreover the refinement increases the resolution of the mesh at every $K=5$ iterations, therefore some facets are subdivided and we need to label the new triangles.

We propose to model the labeling process as a Markov Random Field (MRF) using a simpler data term with respect to \cite{blaha2017semantically}: rather than collecting the likelihood of all the labels for all the images and for each facets, we sample these likelihoods at the vertices locations.
While the geometric term in \cite{blaha2017semantically} considers reasonably handcrafted relationships among the facets and the corresponding labeling, the term we propose estimates the distribution of the normals of the facets belonging to the same class directly from the shape of the current scene. 

Given $\mathcal{F}$ the set of facets and  $\mathcal{L}$ the set of the labels, we aim at assigning a label $l\in \mathcal{L}$ to each facet $f\in\mathcal{F}$, such that we maximize the probability:
\begin{equation}
 P_{\textrm{label}} = \prod_{l\in\mathcal{L}, f\in\mathcal{F}} \left( P_{\textrm{data}}^{lf} \cdot P_{\textrm{norm}}^{lf} \cdot  P_{\textrm{smooth}}^{lf}\right)
\end{equation}

The unary term $P_{\textrm{data}}^{lf}$ describes the evidences of the label $l$ for the facet $f$. 
In principle we need to take into account the whole area of the facet projected in each image. 
However, since our refinement process increases significantly the resolution of the mesh, we simplify the computation of this term only by considering the labels of the pixels where the vertices of the facet are projected and are visible. 
Given the 2D binary masks $M_{2D}^{li}$ of the pixels labeled as $l$ for the point of view of camera $i$, we define for each vertex $v_f$ belonging to $f$:
\begin{equation}
 P_{data}^{lf} = max(\beta, \nicefrac{ \nu(v_f,l)}{3}),
\end{equation}
\begin{equation}
\nu(v_f,l) = \frac{\sum_{i} M_{2D}^{li}(\Pi_i(v_f)) }{\#\textrm{images $v_f$ is visible}},
\end{equation}
where $\beta$ is $0<\beta<1$ prevents $\nu(v_f,l)$ from becoming 0 (we fixed it experimentally to $0.1$), and $\Pi_i(\mathbf{x})$ projects the point $\mathbf{x}$ in the image plane of camera $i$. We divided the term $\nu(v_f,l)$ by 3 such that $0<P_{data}^{lf}<1$.

The unary term $P_{\textrm{norm}}^{lf}$ represents the distribution associated to the class the facet belongs to.
Instead of designing a geometric prior by hand as in \cite{blaha2017semantically} or learning it as in \cite{hane_et_al_09}, we define a method to relate the normals of the  facets belonging to the same class to the scene we are reconstructing.
For each class associated to a label $l$ we estimate the mean normal $\mathbf{m}_l$ and the angle variance $a_l$ with respect to $\mathbf{m}_l$ of all the facet labeled as $l$.
Then we define:
\begin{equation}
 P_{\textrm{norm}}^{lf} = \mu e^{-\frac{\angle(\mathbf{n}_f,\mathbf{m}_l)^2}{2*(a_l)^2}}.
\end{equation}
where $\mu$ weights the importance of $P_{\textrm{norm}}^{lf}$ with respect to the other priors (we fixed $\mu=1.5$)

Finally, we define the binary smoothness term $P_{\textrm{smooth}}^{f_1f_2}$ between two adjacent facets $f_1$ and $f_2$:
\begin{align}
P_{\textrm{smooth}}^{f_1f_2}= 
\begin{cases}
0.2, & \text{if } \mathcal{L}(f_1) \neq \mathcal{L}(f_2)\\
0.8, & \text{if } \mathcal{L}(f_1) = \mathcal{L}(f_2)
\end{cases}
\end{align}
where $\mathcal{L}(f)$ represents the label of facet f; this term penalizes changes in the labeling of $f_1$ and $f_2$, to avoid spurious and noisy labels.

\begin{figure*}[th]
\centering
\setlength{\tabcolsep}{1px}
\begin{tabular}{cc}
{\def\svgwidth{0.5\textwidth}
  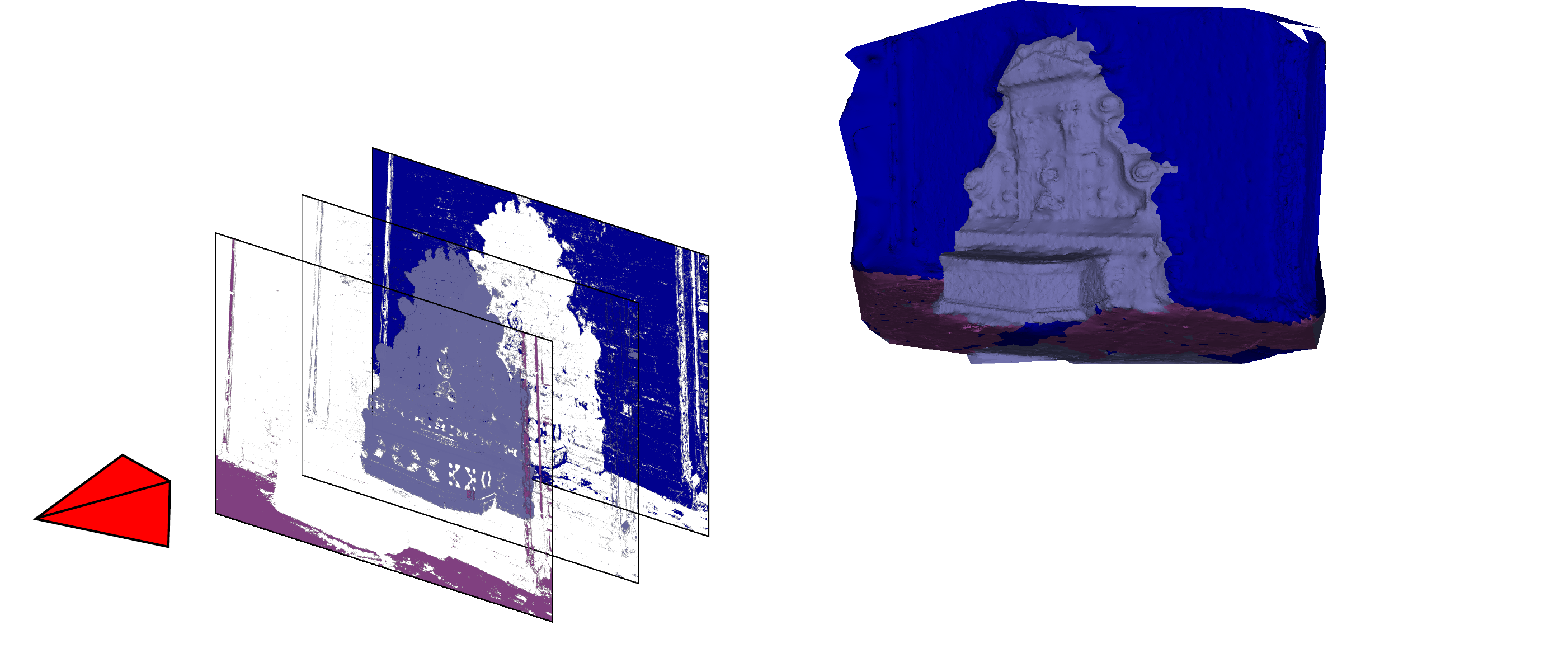}&
{\def\svgwidth{0.5\textwidth}
  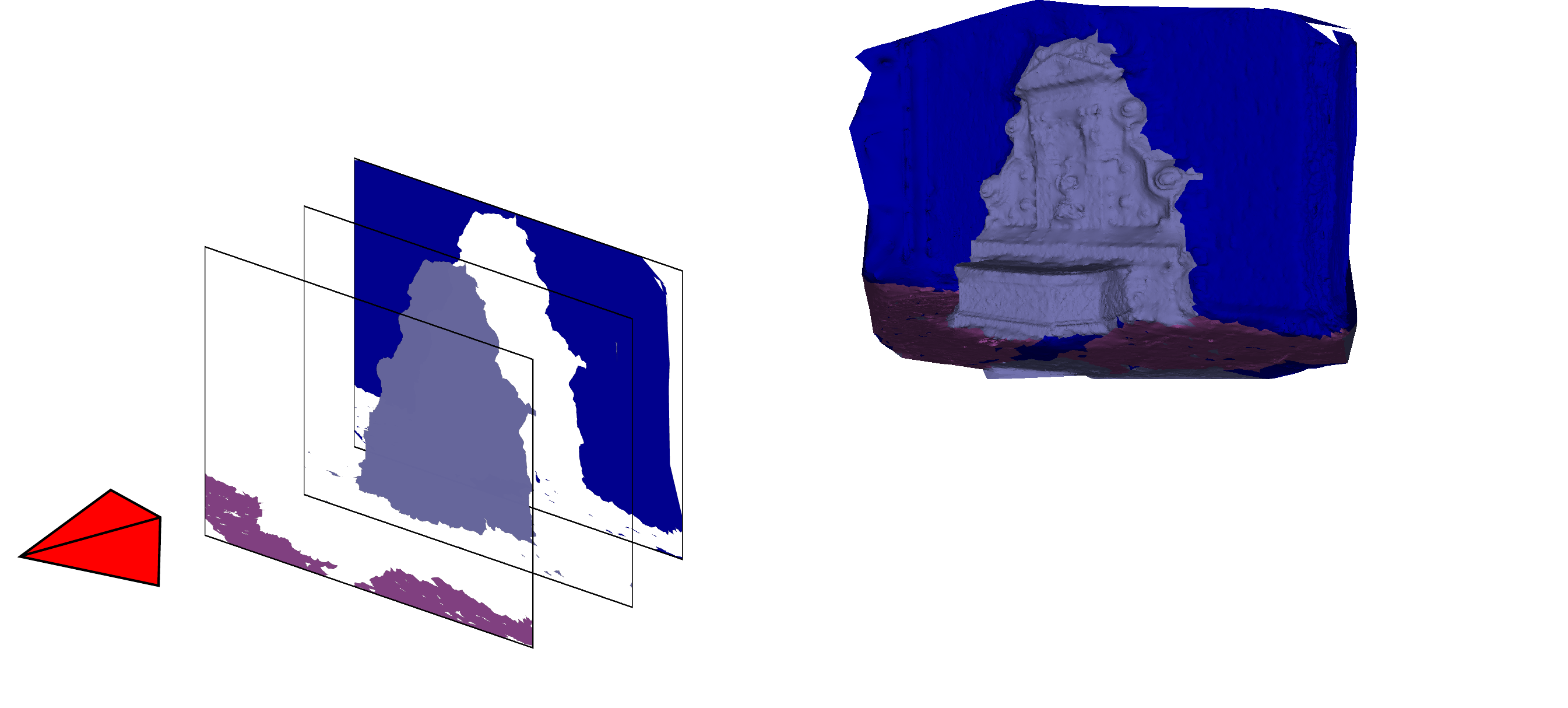}\\
(a)&(b)\\
\end{tabular}
\caption{Masks involved in the semantic mesh refinement: for each class, we compare the masks on the left,  generated from the 3D model, to the masks on the right, that come from the 2D image classification}
\label{fig:semref}
\end{figure*}

\subsection{Semantic Mesh Refinement}
\label{sub:semantic_refinement}
The output of the previous steps is a mesh close to the actual surface of the scene, but it often lacks details.
The most successful method to improve the accuracy of such mesh was proposed by \cite{vu_et_al_2012}.
The idea is to minimize the energy:
\begin{equation}
\label{eq:en}
E = E_{\textrm{photo}} + E_{\textrm{smooth}} ,
\end{equation}
where  $E_{\textrm{photo}}$ is the data term related to the image photo-consistency, and $E_{\textrm{smooth}}$ is a smoothness prior.  

Given a triangular mesh $\mathit{S}$, with $x$ and $\overrightarrow{n}$  a point and the corresponding normal on this mesh, two images $I$ and $J$, and $err_{I, J}(x)$ a function that decreases if the similarity between the patch around the projection of $x$ in $J$ and $I$ increases, then:
\begin{equation}
\label{eq:energy_photo}
  E_{\textrm{photo}} = \sum_{i,j}\int_{\Omega^{\textrm{S}}_{i,j}} err_{I, I_{ij}^{\mathit{S}}}(x_i)\textrm{d}x_i,
\end{equation}
where $I_{ij}^{\mathit{S}}$ is the reprojection of the image from the $j$-th camera in the image $I$ through the mesh $\mathit{S}$ and $\Omega_{i,j}$ represents the domain of the mesh where the projection is defined.
The authors in \cite{vu_et_al_2012}  minimize Eq. \eqref{eq:energy_photo} through gradient descent by moving each vertex $X_i \in \mathbb{R}^3$ of the mesh  according to the gradient:
\begin{align}
\label{eq:gradPhoto}
 \begin{split}
  \frac{\textrm{d}E(\mathit{S})}{\textrm{d}X_i} &=  \int_{\mathit{S}} \phi_i(x) \nabla E_{\textrm{photo}}(x) \textrm{d}x,  = \\
  & - \sum_{i,j} \int_{\Omega^{\textrm{S}}_{i,j}} \phi_i(x)  f_{ij}(x_i) /(\overrightarrow{n}^T \mathbf{d}_i)\overrightarrow{n} \textrm{d}x_i,
 \end{split}
\end{align}
\begin{equation}
f_{ij}(x_i) =\partial_2 err_{I, I_{ij}^{\mathit{S}}}(x_i) DI_j(x_j) D\Pi_j(x) \mathbf{d}_i,
\end{equation}
where $\phi_i(x)$ represents the barycentric coordinates if $x$ is in the triangle containing $X_i$, otherwise $\phi_i(x) = 0$;
$\Pi_j$ is the j-th camera projection, the vector $\mathbf{d}_i$ goes from camera $i$ to point $x$, the operator $D$ represents the derivative and $\partial_2 err_{I, I_{ij}^{\mathit{S}}}(x_i)$ is the derivative of the similarity measure $err_{ij}(x)$ with respect to the second image.

In addition to the photo-consistent term, they minimize the energy $E_{\textrm{smooth}}$ by means of the Laplace-Beltrami operator approximated with the umbrella operator \cite{wardetzky2007discrete}, which moves each vertex in the mean position of its neighbors.

The method presented thus far considers only RGB information and a smoothness prior. 
To leverage the semantic labels estimated in the 2D images and on the 3D mesh, we define an energy function:
\begin{equation}
\label{eq:enNew}
E = E_{\textrm{photo}} + E_{\textrm{sem}} + E_{\textrm{smooth}} ,
\end{equation}
where we minimize $E_{\textrm{photo}}$ and $E_{\textrm{smooth}}$ as in \cite{vu_et_al_2012}, and in the term $E_{\textrm{sem}}$ we exploit the semantic information.

While RGB images contain relatively small noise and, to a certain extent, capture the same color for each point of the scene, when we deal with semantic masks the misclassification strongly depends on the perspective of the images and therefore these masks are not completely consistent among each other. 
For instance, if we have a mask $J$ with a misclassified region $r_m$, even if the current 3D model of the scene is perfectly recovered, the reprojection of $J$ (and in turn of $r_m$) through the surface on the $i$-th will unlikely match the misclassification in the mask estimated for camera $i$.
We assume that the labels that come from image segmentation are noisier and more prone to error than the labels of the 3D mesh, which are estimated from the whole set of image segmentation and corrected with the MRF.
For these reason, differently from the pairwise photometric term $E_{\textrm{photo}}$, we propose a single-view refinement method that compares the semantic mask $I$ with the rendering of the labeled mesh on camera $i$ (Figure \ref{fig:semref}).
By doing so, our refinement affects the borders between the classes in the 3D model and we discard all the wrong classification of the single image segmentation.
 
For each  camera $i$ and for each semantic label $l$ we have a semantic mask $M_{2D}^{li}$  defined as $M_{2D}^{li}=1$ where the label is equal to $l$, and $0$ otherwise (in Figure \ref{fig:semref}(a) the binary masks are depicted in the color to discriminate the classes). 
For the same camera $i$ we also project the visible part of the current 3D mesh classified as $l$, to form the semantic mask projection $M_{3D}^{li}$ (see Figure \ref{fig:semref}(a)).
Given these two masks, for all the cameras $i$ we define:
\begin{equation}
\label{eq:energy_photo}
  E_{\textrm{sem}}^{l} = \sum_{i}\int_{I} {err^{\textrm{sem}}_{M_{2D}^{li}, M_{3D}^{li}}}(x_i)\textrm{d}x_i,
\end{equation}
that we minimize descending the discrete gradient defined over the whole image plane $I$ of the $i$-th camera:
\begin{align}
\label{eq:gradSem}
 \begin{split}
  \frac{\textrm{d}E_{\textrm{sem}}(\mathit{S})}{\textrm{d}X_i} &=  \int_{\mathit{S}} \phi_i(x) \nabla E_{\textrm{sem}}(x) \textrm{d}x  = \\
  & =  - \sum_{i,j} \int_{I} \phi_i(x)  f_{i}(x_i) /(\overrightarrow{n}^T \mathbf{d}_i)\overrightarrow{n} \textrm{d}x_i,
 \end{split}
\end{align}
\begin{equation}
f_{i}(x_i) =\partial_2 {err^{\textrm{sem}}_{M_{2D}^{li}, M_{3D}^{li}}}(x_i) DI_i(x_i) D\Pi_i(x) \mathbf{d}_i.
\end{equation}
Differently from Equation \eqref{eq:gradPhoto}, here we use only a single camera $i$ to compute the gradient.

While the typical error measure adopted for  Equation \eqref{eq:gradPhoto}  is the Zero mean Normalized Cross Correlation (ZNCC), here we adopt a modified version of Sum of Squared Differences (SSD); indeed the semantic masks are binary, therefore no illumination normalization and correction is needed.
The standard SSD gives the same relevance to the two images. Here, instead, we have two semantic masks generated in two deeply different ways: by 2D image segmentation and by labeled mesh rendering.
As stated in Section \ref{subsec:eth},  the mesh labeling is usually more robust and less noisy than the image segmentation. 
To neglect these errors that would induce spurious contributions to the mesh refinement flow, we define the following measure in a window $W$:
\begin{equation}
{err^{\textrm{sem}}_{M_{2D}^{li}, M_{3D}^{li}}} =  \chi \sum_{(\mathbf{x})}^W \left(M_{2D}^{li}(\mathbf{x}) - M_{3D}^{li}(\mathbf{x})\right)^2,
\end{equation}
where $\chi=1$ if the window $W$ defined over $M_{3D}^{li}$ contains at least one pixel belonging to the class mask and one pixel outside the class mask, and $\chi=0$ otherwise. 
This neglects the flow induced by the image segmentation in correspondence of mesh regions with homogeneous labeling.

As in \cite{vu_et_al_2012} we apply a coarse to fine approach that increases the resolution of the mesh after a fixed number of iterations. 
The reasons are twofold: it increase too low-resolution region for the input mesh and it  prevents the refinement to get stuck in local minima and therefore improve the accuracy of the final reconstruction.

On one hand the refinement process changes the shape of the mesh, while on the other hand the coarse to fine approach enhances its resolution.
In both cases the labeling estimated before the refinement could become no more consistent with the refined mesh; for this reason we re-apply the mesh-based labeling  presented in Section \ref{subsec:label} every time we increase the resolution of the mesh.

\section{Experiments}
\label{sec:exp}
To test the effectiveness of the proposed approach we reconstructed three different sequences depicting various scenarios: fountain-p11 from the dataset presented in \cite{strecha2008},  sequence 95 of the KITTI dataset \cite{geiger_et_al12} and a sequence of the DTU dataset \cite{aanaes2016large}. Table \ref{tab:datasetStat} summarizes image resolution, number of frames and number of reconstructed facets for each dataset.
We run the experiments on a laptop with a Intel(R) Core(TM) i7-6700HQ CPU at  2.60GHz, 16GB of RAM and  a GeForce GTX 960M.

\begin{table}[t]
\normalsize
\centering
\setlength{\tabcolsep}{5px}
  \caption{Resolutions and output statistic for each dataset we used.}
  \label{tab:datasetStat}
\begin{tabular}{lccccc}
&num.& image &num.\\
&cameras& resolution&facets \\
\hline
fountain-p11	&11 	& 3072x2048		& 1.9M\\
KITTI 95		&512 	& 1242x375		& 2.6M\\
DTU 15			&49 	& 1600x1200 	& 0.6M\\
\end{tabular}
\end{table}

One of the inputs of the proposed algorithm is the semantic segmentation of the images.
For the fountain-p11 and the DTU sequence we manually annotated a few images and we trained a Multiboost classifier \cite{benbouzid2012multiboost} on them; since the KITTI sequence is more challenging, we used ReSeg \cite{visin2016reseg}, a Recurrent Neural Network based model trained on the Cityscapes dataset \cite{cordts2016cityscapes}.
The points adopted in our modified semantic version of \cite{romanoni15b} are a combination of Structure from Motion points \cite{openMVG}, semi-global stereo matching \cite{hirschmuller2008stereo} and plane sweeping \cite{hane2014real}. 

We evaluate the accuracy of the 3D reconstruction with the method described by \cite{strecha2008}: we consider the most significant image or images and, from the same point of view, we compute and compare the depth map generated with the ground truth and the reconstructed 3D model. 
For the fountain-p11 and DTU dataset we choose one image that captures the whole scene, and for the KITTI sequence we computed the depth maps from five images spread along the path.
In Table \ref{tab:refinement} we illustrate the reconstruction errors (expressed in mm) of our method compared with the modified \cite{romanoni15b}, the refinement in \cite{vu_et_al_2012} and the joint semantic and photometric refinement presented in \cite{blaha2017semantically}, applied to our labeled mesh: for all the three datasets our method improve the reconstruction error.
This proves that the semantic information coupled with the photo-metric term, improves the convergence of the refinement algorithm.

To evaluate the quality of semantic labeling we project the labeled image into the same cameras we adopted to compute the depth map, and we compare them against manually annotate images. 
We compare against the 3D methods \cite{romanoni15b}, \cite{vu_et_al_2012} and \cite{blaha2017semantically} and the 2D semantic segmentation from \cite{benbouzid2012multiboost} and \cite{visin2016reseg}, inputs of our algorithm.
We show the results in Table \ref{tab:segm}: we listed several classical metric adopted in classification problems: accuracy, recall, F-score and precision.
Except for the recall of the KITTI dataset, our algorithm achieves the best performances in all the datasets for each metric.
This proves that the relabeling we adopted is effective and it especially regularize the labels where the noise affects the input semantic segmentation.
In the KITTI dataset, where the initial image segmentations contains less noise with respect to the other dataset, the results of our refinement and \cite{blaha2017semantically} are very close.


\begin{table}[t]
\centering
 \setlength{\tabcolsep}{5px}
  \caption{Reconstruction Accuracy measured with Mean Absolute Error and expressed in mm.}
  \label{tab:refinement}
    \begin{tabular}{lcccc}
    &\cite{romanoni15b}&\cite{vu_et_al_2012}&\cite{blaha2017semantically}& Proposed\\
    \hline
     fountain-p11&12.7&9.2&8.6&\textbf{8.5} \\
     KITTI 95&46.7 & 32.8 & \textbf{32.7} &\textbf{32.7}\\
     DTU 15 &2.64 & 2.47&2.57& \textbf{2.40}\\
    \end{tabular}
\end{table}

\begin{table*}[tp]
\centering
 \setlength{\tabcolsep}{15px}
  \caption{Segmentation statistics.}
  \label{tab:segm}
    \begin{tabular}{llcccc}
                   & &                             accuracy  &  recall  &  F-score   & precision\\
    \hline
\multirow{5}{*}{Fountain}                                       
& Multiboost \cite{benbouzid2012multiboost}   &  0.9144   &   0.8495  &  0.8462   &  0.8594\\
& Semantic \cite{romanoni15b}                 &  0.9425   &   0.8318  &  0.8592   &  0.9145\\
&\cite{vu_et_al_2012}                         &  0.9400   &   0.8256  &  0.8533   &  0.9095\\
&\cite{blaha2017semantically}                 &  {0.9532}   &   {0.8679}  &  {0.8923}   &  {0.9295}\\
&Proposed                                     &  \textbf{0.9571}   &   \textbf{0.8755}  &  \textbf{0.9003}   &  \textbf{0.9385}\\
\hline
\multirow{4}{*}{DTU}
& Multiboost \cite{benbouzid2012multiboost}   &  0.9043  &    0.7230  &  0.6991   &  0.6837\\
& Semantic \cite{romanoni15b}                 &  0.9204  &    0.6753  &  0.6837   &  0.7241\\
&\cite{vu_et_al_2012}                         &  0.9226  &    0.6617  &  0.6782   &  0.7311\\
&\cite{blaha2017semantically}                 &  0.9551  &    0.7843  &  0.7920   &  0.8242\\
&Proposed                                     &  \textbf{0.9561} &    \textbf{0.7935}  &  \textbf{0.8000}   &  \textbf{0.8329}\\
\hline
\multirow{4}{*}{KITTI 95}
&ReSeg \cite{visin2016reseg}                  & 0.9700  &    0.9117  &  0.9092  &  0.9140\\
& Semantic \cite{romanoni15b}                 & 0.9668  &    0.9093  &  0.8968  &  0.8906\\
&\cite{vu_et_al_2012}                         & 0.9672  &    0.9107  &  0.8984  &  0.8922\\
&\cite{blaha2017semantically}                                      & \textbf{0.9709}  &    \textbf{0.9246}  &  0.9107  &  0.9084\\
&Proposed                                     & \textbf{0.9709}  &    0.9241  &  \textbf{0.9109}  &  \textbf{0.9089}
    \end{tabular}
\end{table*}

\begin{figure*}[tp]
\centering
\setlength{\tabcolsep}{1px}
\begin{tabular}{ccccc}
labelled image&Semantic \cite{romanoni15b} &\cite{vu_et_al_2012}&\cite{blaha2017semantically}  &Proposed\\
\hline
\includegraphics[width=0.2\textwidth]{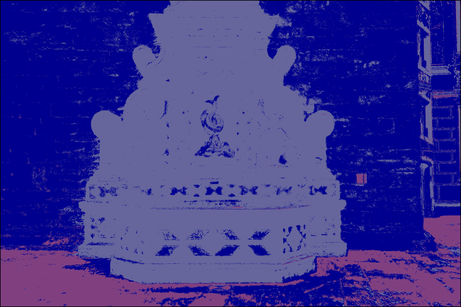}&
\includegraphics[width=0.2\textwidth]{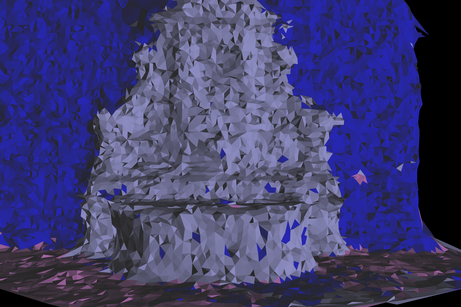}&
\includegraphics[width=0.2\textwidth]{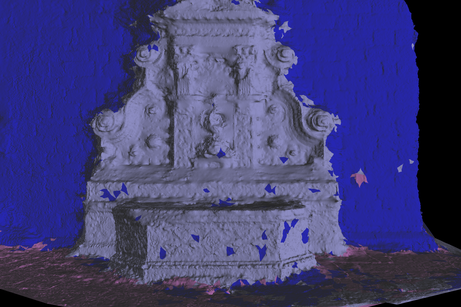}&
\includegraphics[width=0.2\textwidth]{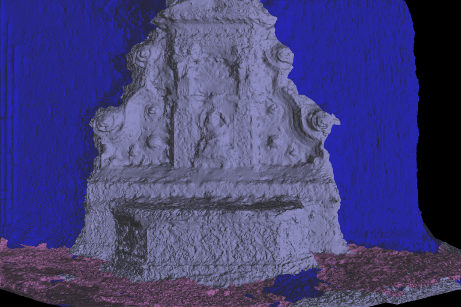}&
\includegraphics[width=0.2\textwidth]{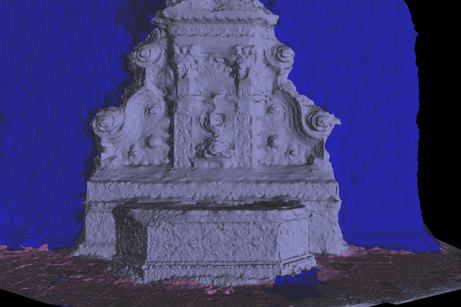}\\
\multicolumn{4}{c}{fountain-p11}\\
\includegraphics[width=0.2\textwidth]{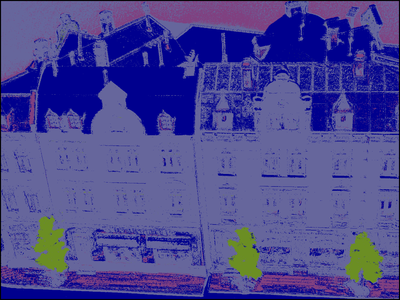}&
\includegraphics[width=0.2\textwidth]{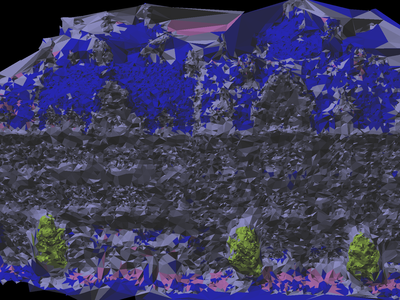}&
\includegraphics[width=0.2\textwidth]{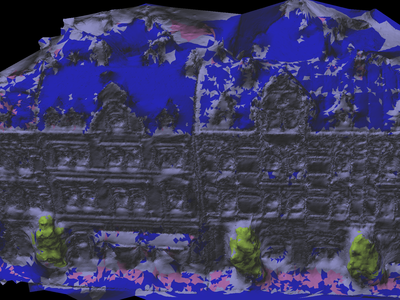}&
\includegraphics[width=0.2\textwidth]{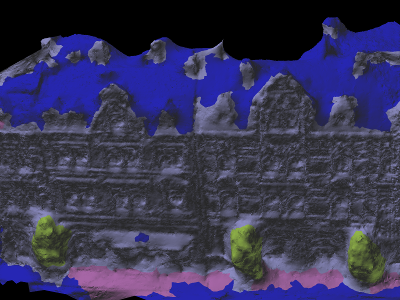}&
\includegraphics[width=0.2\textwidth]{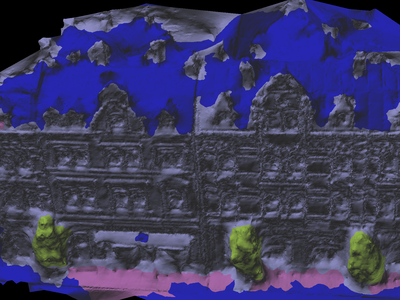}\\
\multicolumn{4}{c}{DTU sequence 15}\\
\includegraphics[width=0.2\textwidth,height=0.125\textwidth]{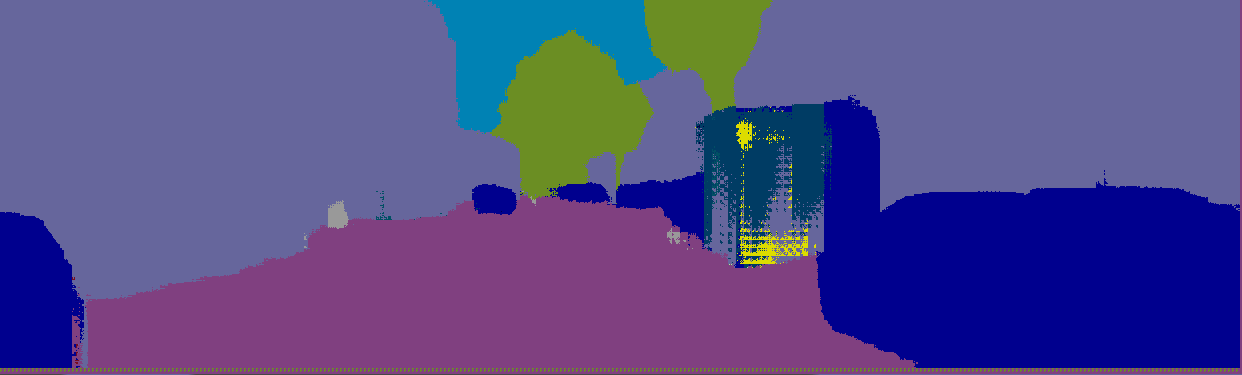}&
\includegraphics[width=0.2\textwidth,height=0.125\textwidth]{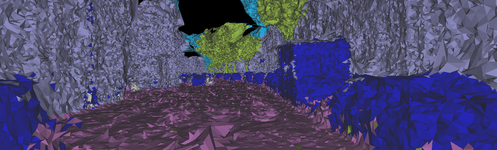}&
\includegraphics[width=0.2\textwidth,height=0.125\textwidth]{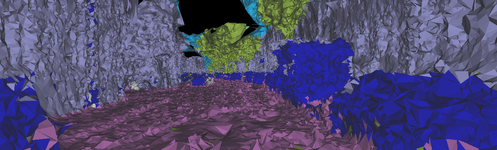}&
\includegraphics[width=0.2\textwidth,height=0.125\textwidth]{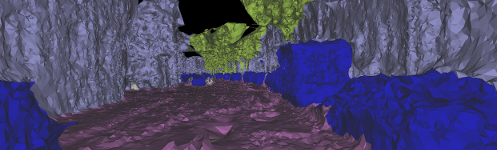}&
\includegraphics[width=0.2\textwidth,height=0.125\textwidth]{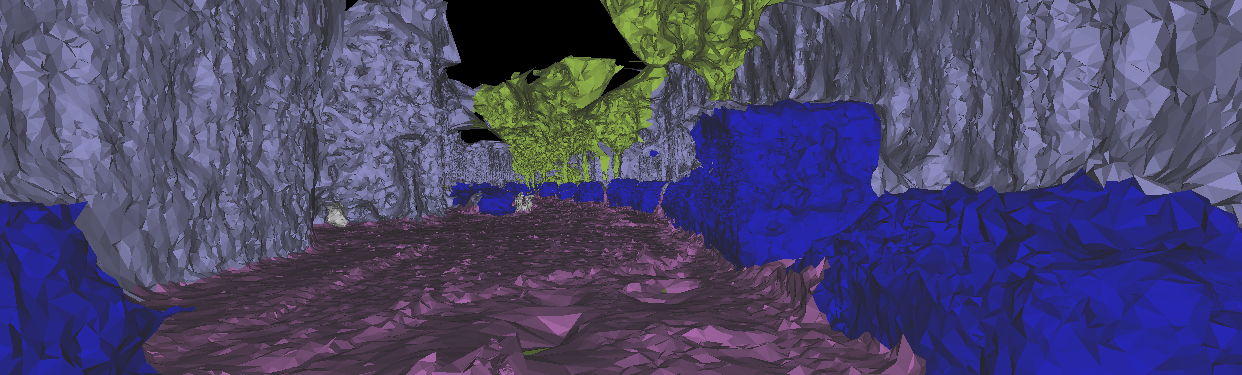}\\
\multicolumn{4}{c}{KITTI sequence 95}\\
\end{tabular}
\caption{Results on fountain-p11, DTU and KITTI datasets}
\label{fig:results}
\end{figure*}


\begin{figure*}[tp]
\centering
\setlength{\tabcolsep}{1px}
\begin{tabular}{ccc}
Semantic \cite{romanoni15b} &\cite{vu_et_al_2012}  &Proposed\\
\hline
\includegraphics[width=0.32\textwidth]{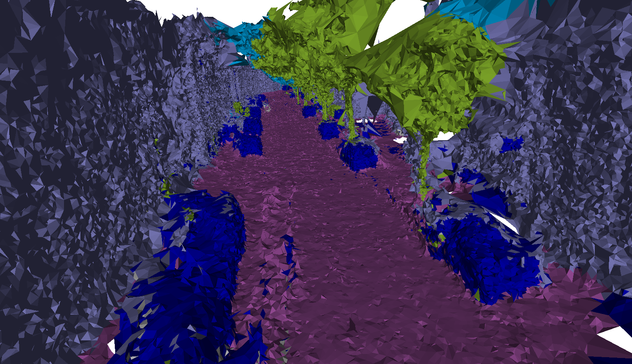}&
\includegraphics[width=0.32\textwidth]{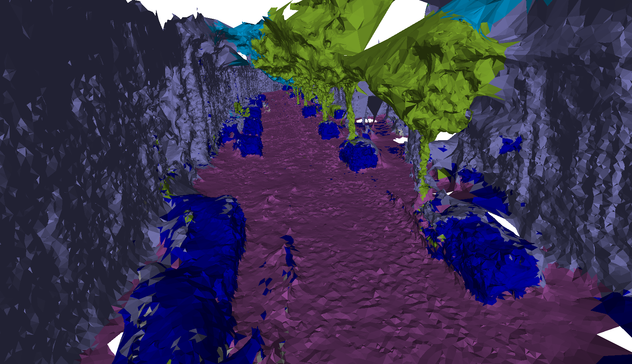}&
\includegraphics[width=0.32\textwidth]{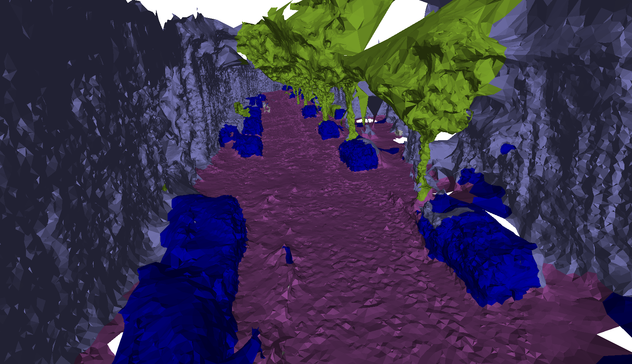}\\
\end{tabular}
\caption{A wide view of the KITTI reconstruction}
\label{fig:kitti}
\end{figure*}

\begin{figure*}[tp]
\centering
\setlength{\tabcolsep}{1px}
\begin{tabular}{ccc}
  \begin{adjustbox}{valign=c}
    \begin{tabular}{@{}c@{}}
    \includegraphics[width=0.3\textwidth]{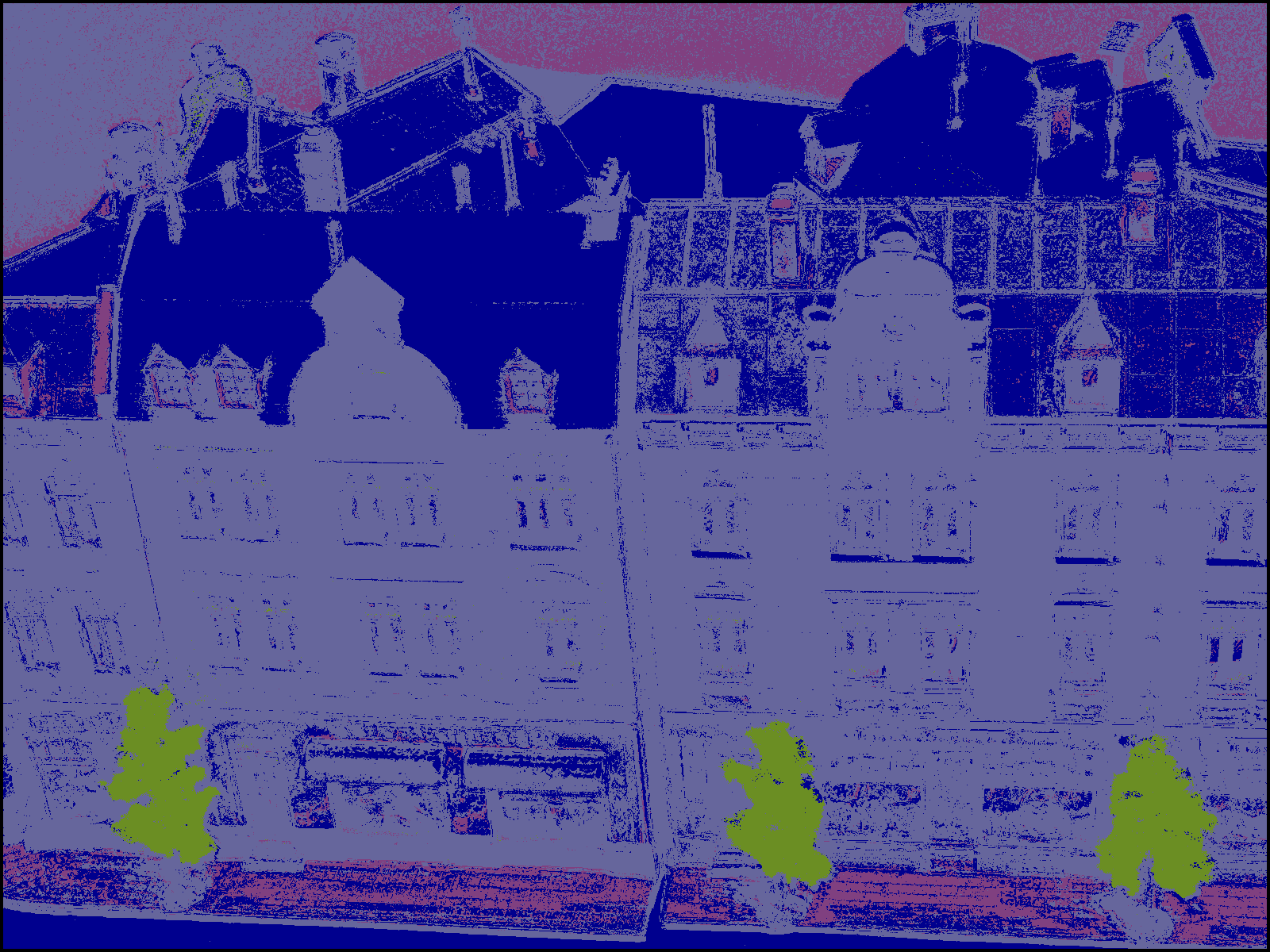}\\
    (a)
    \end{tabular}
  \end{adjustbox}
  &
  \begin{adjustbox}{valign=c}
    \begin{tabular}{@{}c@{}}
     	\includegraphics[width=0.3\textwidth]{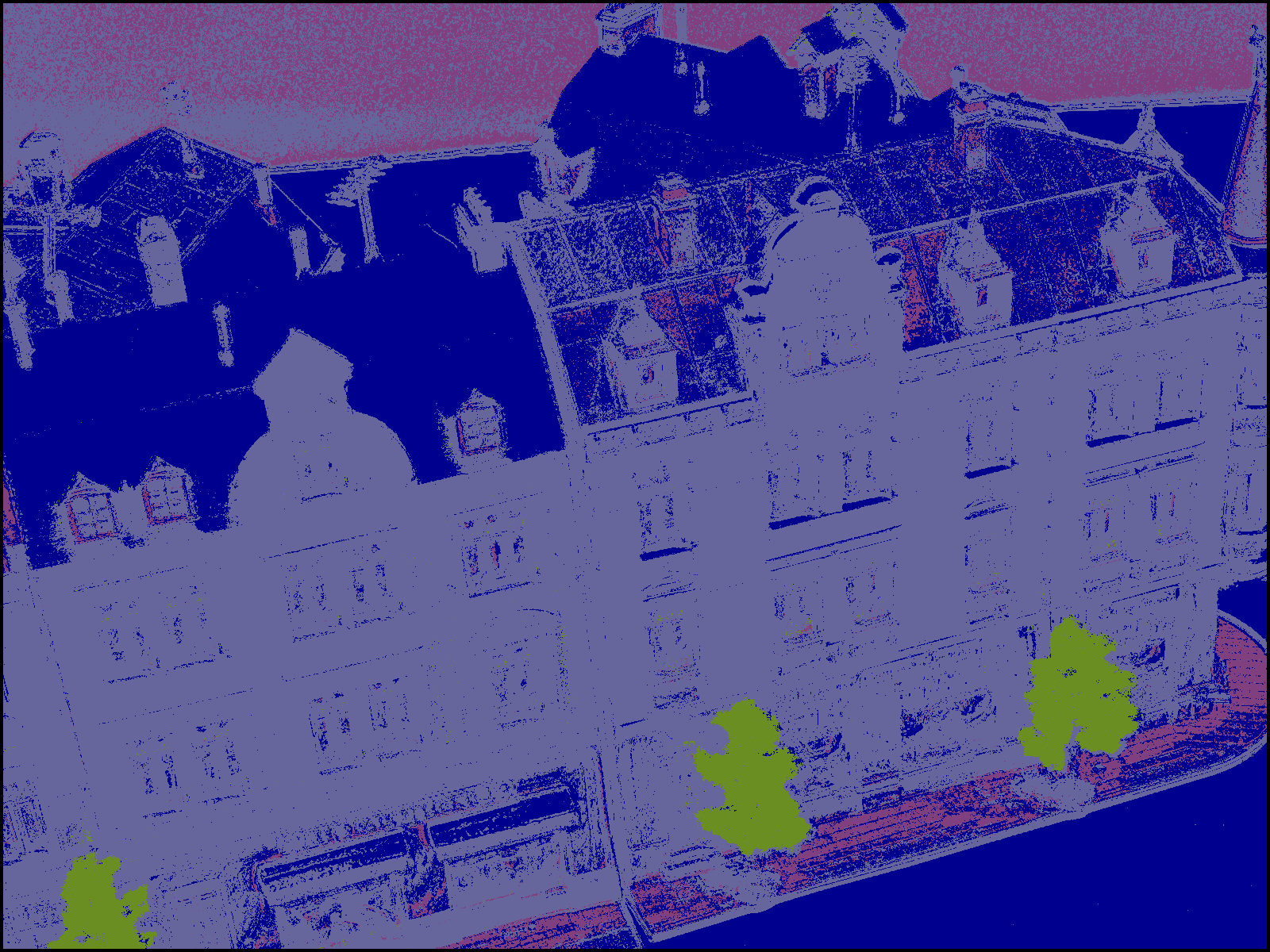}\\(b)\\[1ex]
      \includegraphics[width=0.3\textwidth]{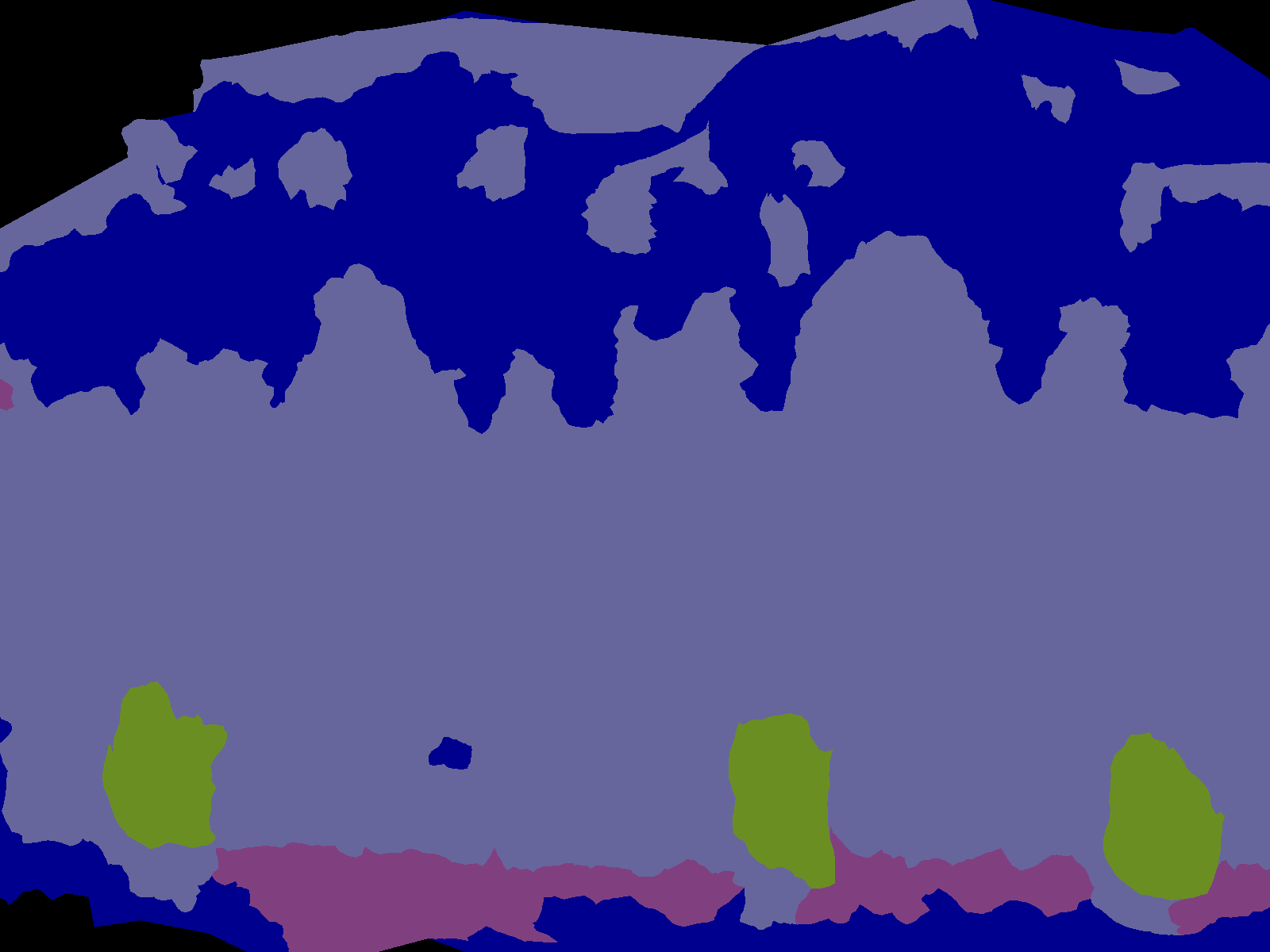}\\(d)
    \end{tabular}
  \end{adjustbox}
  &
  \begin{adjustbox}{valign=c}
    \begin{tabular}{@{}c@{}}
       \includegraphics[width=0.3\textwidth]{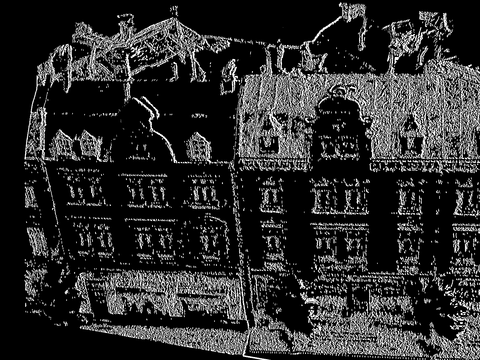}\\(c)\\[1ex]
       \includegraphics[width=0.3\textwidth]{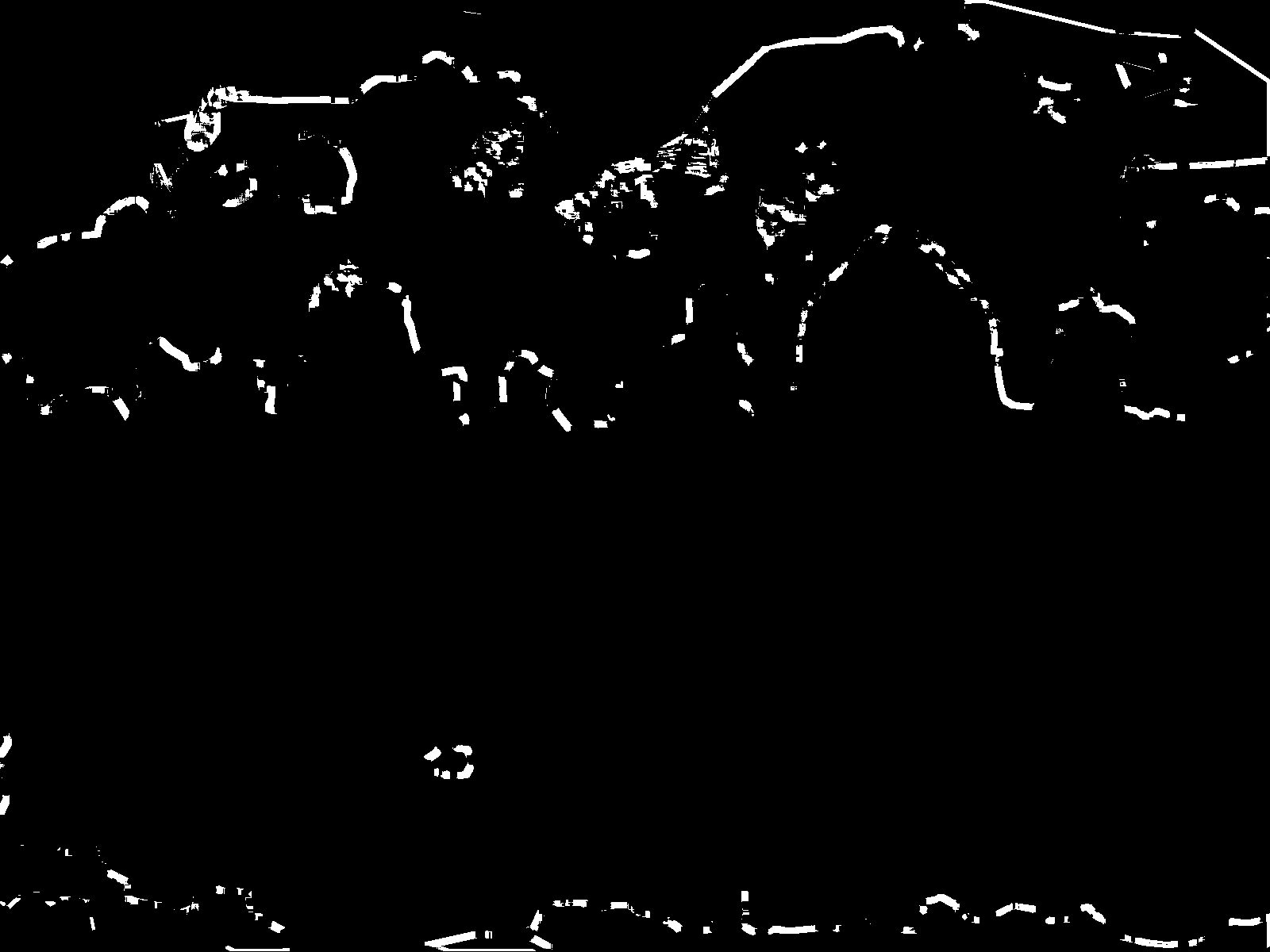}\\(e)
    \end{tabular}
  \end{adjustbox}\\
  labels considered in the first term of &  labels considered in the  second term of &  non-zero \\
  the similarity measure&  the similarity measure&gradients \\[1ex]
\end{tabular}
\caption{Comparison of the gradients computed by the method presented in \cite{blaha2017semantically} (top) and the method we presented in this paper (bottom). In the first and second columns we show the two terms compared by the similarity measure; in the third column the resulting gradients. Notice that our method uses a single point of view.}
\label{fig:twoviews}
\end{figure*}

\subsection{Comparison with two view semantic mesh refinement}
\label{subsec:twoviews}
The method we presented in this paper refines the mesh accordingly to both the photometric and semantic information, in a similar yet quite different way to the very recent work appeared in \cite{blaha2017semantically}.
For each label $l$ defined in the image classifier, both methods compare two masks containing the pixels classified with the label $l$, and modify the shape of the mesh to minimize the reprojection errors of the second mask though the mesh into the first mask.

While in \cite{blaha2017semantically} both the first (Figure \ref{fig:twoviews}(a)) and the second masks  (Figure \ref{fig:twoviews}(b)) are the outputs of the 2D image classifier, in this paper we propose a single-view method that compares the masks from camera $i$ (Figure \ref{fig:twoviews}(a)) with the mask rendered from the 3D labeled mesh to the same point of view of camera $i$ (bottom of Figure \ref{fig:twoviews}(d)).

To verify that, as stated in Section \ref{subsec:eth}, our method is robust to the noise and errors that often affect the image segmentations, we implemented the method \cite{blaha2017semantically}. We applied it to the fa\c{c}ade masks obtained from Figure \ref{fig:twoviews}(a) and Figure \ref{fig:twoviews}(b); Figure \ref{fig:twoviews}(c) shows the mask of non-zero gradients.
On the other hand, in Figure \ref{fig:twoviews}(e), we compute the gradients with our method by comparing the masks from Figure \ref{fig:twoviews}(a) and the rendered mesh in Figure \ref{fig:twoviews}(d).


Figure \ref{fig:twoviews}(e) shows that the method in \cite{blaha2017semantically} cumulate the noise from Figure \ref{fig:twoviews}(a) and Figure \ref{fig:twoviews}(b); all the contributions outside the neighborhood of the real class borders are the consequences of misclassification in the two compared masks, therefore they evolve the mesh incoherently. 
These errors cumulate cross all the pairwise comparison since the classification errors are different for each view and the pairwise contributions corresponding to their location in general are not mutually compensated along the sequence.
Even if the smoothing term of the refinement diminish these errors, they affect the final reconstruction.
As a further proof, in Table \ref{tab:refinement} and in Table \ref{tab:segm}  our approach overcome the one in \cite{blaha2017semantically} especially in the DTU dataset, where the segmented images are very noisy.

Instead, our method computes a cleaner gradient flow (Figure \ref{fig:twoviews}(e)) thanks to the comparison with the mask rendered from the labeled mesh, that, after the MRF labeling, is robust to noise and errors.

\section{Conclusions and Future works}
\label{sec:concl}
In this paper we presented a novel method to refine a semantically annotated mesh through single-view variational energy minimization coupled with the photo-metric term. 
We also propose to update the labels as the shape of the reconstruction is modified, in particular our contribution in this case is a MRF formulation that takes into account class-specific normal prior that is estimated from the existing annotated mesh instead of the handcrafted or learned priors proposed in the literature.

The refinement algorithm proposed in this paper could be further extended by adding  geometric priors or we could investigate how it can enforce the convergence in challenging dataset, e.g., when the texture is almost flat.
We also plan to evaluate how the accuracy of the initial mesh could affect the final reconstruction with or without the semantic refinement term.


\section*{Acknowledgments}
{\small
This work has been supported by the ``Interaction between Driver Road Infrastructure Vehicle and Environment (I.DRIVE)'' Inter-department Laboratory from Politecnico di Milano, and the ``Cloud4Drones'' project founded by EIT Digital. We thank Nvidia who has kindly supported our research through the Hardware Grant Program. }
{\small
\bibliographystyle{ieee}
\bibliography{biblioTotal}

\begin{thebibliography}{10}\itemsep=-1pt

\bibitem{aanaes2016large}
H.~Aan{\ae}s, R.~R. Jensen, G.~Vogiatzis, E.~Tola, and A.~B. Dahl.
\newblock Large-scale data for multiple-view stereopsis.
\newblock {\em International Journal of Computer Vision}, pages 1--16, 2016.

\bibitem{benbouzid2012multiboost}
D.~Benbouzid, R.~Busa-Fekete, N.~Casagrande, F.-D. Collin, and B.~K{\'e}gl.
\newblock Multiboost: a multi-purpose boosting package.
\newblock {\em Journal of Machine Learning Research}, 13(Mar):549--553, 2012.

\bibitem{blaha2017semantically}
M.~Blaha, M.~Rothermel, M.~R. Oswald, T.~Sattler, A.~Richard, J.~D. Wegner,
  M.~Pollefeys, and K.~Schindler.
\newblock Semantically informed multiview surface refinement.
\newblock {\em arXiv preprint arXiv:1706.08336}, 2017.

\bibitem{blaha2016large}
M.~Blaha, C.~Vogel, A.~Richard, J.~D. Wegner, T.~Pock, and K.~Schindler.
\newblock Large-scale semantic 3d reconstruction: an adaptive multi-resolution
  model for multi-class volumetric labeling.
\newblock In {\em Proceedings of the IEEE Conference on Computer Vision and
  Pattern Recognition}, pages 3176--3184, 2016.

\bibitem{cabezas2015semantically}
R.~Cabezas, J.~Straub, and J.~W. Fisher.
\newblock Semantically-aware aerial reconstruction from multi-modal data.
\newblock In {\em Proceedings of the IEEE International Conference on Computer
  Vision}, pages 2156--2164, 2015.

\bibitem{chen2016deeplab}
L.-C. Chen, G.~Papandreou, I.~Kokkinos, K.~Murphy, and A.~L. Yuille.
\newblock Deeplab: Semantic image segmentation with deep convolutional nets,
  atrous convolution, and fully connected crfs.
\newblock {\em arXiv preprint arXiv:1606.00915}, 2016.

\bibitem{cherabier2016multi}
I.~Cherabier, C.~H{\"a}ne, M.~R. Oswald, and M.~Pollefeys.
\newblock Multi-label semantic 3d reconstruction using voxel blocks.
\newblock In {\em 3D Vision (3DV), 2016 Fourth International Conference on},
  pages 601--610. IEEE, 2016.

\bibitem{cordts2016cityscapes}
M.~Cordts, M.~Omran, S.~Ramos, T.~Rehfeld, M.~Enzweiler, R.~Benenson,
  U.~Franke, S.~Roth, and B.~Schiele.
\newblock The cityscapes dataset for semantic urban scene understanding.
\newblock In {\em Proceedings of the IEEE Conference on Computer Vision and
  Pattern Recognition}, pages 3213--3223, 2016.

\bibitem{delaunoy2014photometric}
A.~Delaunoy and M.~Pollefeys.
\newblock Photometric bundle adjustment for dense multi-view 3d modeling.
\newblock In {\em Computer Vision and Pattern Recognition (CVPR), 2014 IEEE
  Conference on}, pages 1486--1493. IEEE, 2014.

\bibitem{delaunoy_et_al_08}
A.~Delaunoy, E.~Prados, P.~G.~I. Pirac{\'e}s, J.-P. Pons, and P.~Sturm.
\newblock Minimizing the multi-view stereo reprojection error for triangular
  surface meshes.
\newblock In {\em BMVC 2008-British Machine Vision Conference}, pages 1--10.
  BMVA, 2008.

\bibitem{gargallo2007minimizing}
P.~Gargallo, E.~Prados, and P.~Sturm.
\newblock Minimizing the reprojection error in surface reconstruction from
  images.
\newblock In {\em Computer Vision, 2007. ICCV 2007. IEEE 11th International
  Conference on}, pages 1--8. IEEE, 2007.

\bibitem{geiger_et_al12}
A.~Geiger, P.~Lenz, and R.~Urtasun.
\newblock Are we ready for autonomous driving? the kitti vision benchmark
  suite.
\newblock In {\em Computer Vision and Pattern Recognition (CVPR), 2012 IEEE
  Conference on}, pages 3354--3361. IEEE, 2012.

\bibitem{hane2014real}
C.~H{\"a}ne, L.~Heng, G.~H. Lee, A.~Sizov, and M.~Pollefeys.
\newblock Real-time direct dense matching on fisheye images using
  plane-sweeping stereo.
\newblock In {\em 3D Vision (3DV), 2014 2nd International Conference on},
  volume~1, pages 57--64. IEEE, 2014.

\bibitem{hane2016overview}
C.~H{\"a}ne and M.~Pollefeys.
\newblock An overview of recent progress in volumetric semantic 3d
  reconstruction.
\newblock In {\em Pattern Recognition (ICPR), 2016 23rd International
  Conference on}, pages 3294--3307. IEEE, 2016.

\bibitem{hane2013joint}
C.~Hane, C.~Zach, A.~Cohen, R.~Angst, and M.~Pollefeys.
\newblock Joint 3d scene reconstruction and class segmentation.
\newblock In {\em Computer Vision and Pattern Recognition (CVPR), 2013 IEEE
  Conference on}, pages 97--104. IEEE, 2013.

\bibitem{hane_et_al_09}
C.~Hane, C.~Zach, A.~Cohen, R.~Angst, and M.~Pollefeys.
\newblock Joint 3d scene reconstruction and class segmentation.
\newblock In {\em Computer Vision and Pattern Recognition (CVPR), 2013 IEEE
  Conference on}, pages 97--104. IEEE, 2013.

\bibitem{hirschmuller2008stereo}
H.~Hirschmuller.
\newblock Stereo processing by semiglobal matching and mutual information.
\newblock {\em IEEE Transactions on pattern analysis and machine intelligence},
  30(2):328--341, 2008.

\bibitem{kundu2014joint}
A.~Kundu, Y.~Li, F.~Dellaert, F.~Li, and J.~M. Rehg.
\newblock Joint semantic segmentation and 3d reconstruction from monocular
  video.
\newblock In {\em European Conference on Computer Vision}, pages 703--718.
  Springer, 2014.

\bibitem{li2016efficient}
S.~Li, S.~Y. Siu, T.~Fang, and L.~Quan.
\newblock Efficient multi-view surface refinement with adaptive resolution
  control.
\newblock In {\em European Conference on Computer Vision}, pages 349--364.
  Springer, 2016.

\bibitem{litvinov_lhiuller14}
V.~Litvinov and M.~Lhuillier.
\newblock Incremental solid modeling from sparse structure-from-motion data
  with improved visual artifacts removal.
\newblock In {\em International Conference on Pattern Recognition (ICPR)},
  2014.

\bibitem{openMVG}
P.~Moulon, P.~Monasse, R.~Marlet, and Others.
\newblock Openmvg. an open multiple view geometry library.
\newblock \url{https://github.com/openMVG/openMVG}.

\bibitem{newcombe2011dtam}
R.~A. Newcombe, S.~J. Lovegrove, and A.~J. Davison.
\newblock Dtam: Dense tracking and mapping in real-time.
\newblock In {\em Computer Vision (ICCV), 2011 IEEE International Conference
  on}, pages 2320--2327. IEEE, 2011.

\bibitem{pollefeys_et_al_08}
M.~Pollefeys, D.~Nist{\'e}r, J.-M. Frahm, A.~Akbarzadeh, P.~Mordohai, B.~Clipp,
  C.~Engels, D.~Gallup, S.-J. Kim, P.~Merrell, et~al.
\newblock Detailed real-time urban 3d reconstruction from video.
\newblock {\em International Journal of Computer Vision}, 78(2-3):143--167,
  2008.

\bibitem{pons2007multi}
J.-P. Pons, R.~Keriven, and O.~Faugeras.
\newblock Multi-view stereo reconstruction and scene flow estimation with a
  global image-based matching score.
\newblock {\em International Journal of Computer Vision}, 72(2):179--193, 2007.

\bibitem{romanoni16}
A.~Romanoni, A.~Delaunoy, M.~Pollefeys, and M.~Matteucci.
\newblock Automatic 3d reconstruction of manifold meshes via delaunay
  triangulation and mesh sweeping.
\newblock In {\em Winter Conference on Applications of Computer Vision (WACV)}.
  IEEE, 2016.

\bibitem{romanoni15b}
A.~Romanoni and M.~Matteucci.
\newblock Incremental reconstruction of urban environments by edge-points
  delaunay triangulation.
\newblock In {\em Intelligent Robots and Systems (IROS), 2015 IEEE/RSJ
  International Conference on}, pages 4473--4479. IEEE, 2015.

\bibitem{rouhani2017semantic}
M.~Rouhani, F.~Lafarge, and P.~Alliez.
\newblock Semantic segmentation of 3d textured meshes for urban scene analysis.
\newblock {\em ISPRS Journal of Photogrammetry and Remote Sensing},
  123:124--139, 2017.

\bibitem{savinov2015discrete}
N.~Savinov, L.~Ladicky, C.~Hane, and M.~Pollefeys.
\newblock Discrete optimization of ray potentials for semantic 3d
  reconstruction.
\newblock In {\em Computer Vision and Pattern Recognition (CVPR), 2015 IEEE
  Conference on}, pages 5511--5518. IEEE, 2015.

\bibitem{schops20153d}
T.~Schops, T.~Sattler, C.~Hane, and M.~Pollefeys.
\newblock 3d modeling on the go: Interactive 3d reconstruction of large-scale
  scenes on mobile devices.
\newblock In {\em 3D Vision (3DV), 2015 International Conference on}, pages
  291--299. IEEE, 2015.

\bibitem{sengupta2013urban}
S.~Sengupta, E.~Greveson, A.~Shahrokni, and P.~H. Torr.
\newblock Urban 3d semantic modelling using stereo vision.
\newblock In {\em Robotics and Automation (ICRA), 2013 IEEE International
  Conference on}, pages 580--585. IEEE, 2013.

\bibitem{steinbrucker2014volumetric}
F.~Steinbrucker, J.~Sturm, and D.~Cremers.
\newblock Volumetric 3d mapping in real-time on a cpu.
\newblock In {\em Robotics and Automation (ICRA), 2014 IEEE International
  Conference on}, pages 2021--2028. IEEE, 2014.

\bibitem{strecha2008}
C.~Strecha, W.~von Hansen, L.~Van~Gool, P.~Fua, and U.~Thoennessen.
\newblock On benchmarking camera calibration and multi-view stereo for high
  resolution imagery.
\newblock In {\em Computer Vision and Pattern Recognition, 2008. CVPR 2008.
  IEEE Conference on}, pages 1--8. IEEE, 2008.

\bibitem{tola12}
E.~Tola, C.~Strecha, and P.~Fua.
\newblock Efficient large-scale multi-view stereo for ultra high-resolution
  image sets.
\newblock {\em Machine Vision and Applications}, 23(5):903--920, 2012.

\bibitem{valentin2013mesh}
J.~P. Valentin, S.~Sengupta, J.~Warrell, A.~Shahrokni, and P.~H. Torr.
\newblock Mesh based semantic modelling for indoor and outdoor scenes.
\newblock In {\em Proceedings of the IEEE Conference on Computer Vision and
  Pattern Recognition}, pages 2067--2074, 2013.

\bibitem{verdie2015lod}
Y.~Verdie, F.~Lafarge, and P.~Alliez.
\newblock Lod generation for urban scenes.
\newblock Technical report, Association for Computing Machinery, 2015.

\bibitem{visin2016reseg}
F.~Visin, M.~Ciccone, A.~Romero, K.~Kastner, K.~Cho, Y.~Bengio, M.~Matteucci,
  and A.~Courville.
\newblock Reseg: A recurrent neural network-based model for semantic
  segmentation.
\newblock In {\em Proceedings of the IEEE Conference on Computer Vision and
  Pattern Recognition Workshops}, pages 41--48, 2016.

\bibitem{vogiatzis2005multi}
G.~Vogiatzis, P.~H. Torr, and R.~Cipolla.
\newblock Multi-view stereo via volumetric graph-cuts.
\newblock In {\em Computer Vision and Pattern Recognition, 2005. CVPR 2005.
  IEEE Computer Society Conference on}, volume~2, pages 391--398. IEEE, 2005.

\bibitem{vu2011large}
H.~H. Vu.
\newblock {\em Large-scale and high-quality multi-view stereo}.
\newblock PhD thesis, Paris Est, 2011.

\bibitem{vu_et_al_2012}
H.~H. Vu, P.~Labatut, J.-P. Pons, and R.~Keriven.
\newblock High accuracy and visibility-consistent dense multiview stereo.
\newblock {\em Pattern Analysis and Machine Intelligence, IEEE Transactions
  on}, 34(5):889--901, 2012.

\bibitem{wardetzky2007discrete}
M.~Wardetzky, S.~Mathur, F.~K{\"a}lberer, and E.~Grinspun.
\newblock Discrete laplace operators: no free lunch.
\newblock In {\em Symposium on Geometry processing}, pages 33--37, 2007.

\bibitem{yezzi2003stereoscopic}
A.~Yezzi and S.~Soatto.
\newblock Stereoscopic segmentation.
\newblock {\em International Journal of Computer Vision}, 53(1):31--43, 2003.

\bibitem{zheng2015conditional}
S.~Zheng, S.~Jayasumana, B.~Romera-Paredes, V.~Vineet, Z.~Su, D.~Du, C.~Huang,
  and P.~H. Torr.
\newblock Conditional random fields as recurrent neural networks.
\newblock In {\em Proceedings of the IEEE International Conference on Computer
  Vision}, pages 1529--1537, 2015.

\end{thebibliography}
}

\end{document}